\documentclass[11pt]{article}

\pdfoutput=1

\usepackage{setspace}
\usepackage{appendix}
\usepackage{xfrac}
\usepackage{amsmath, amssymb,amsfonts}
\usepackage[figuresright]{rotating}
\usepackage{amsmath,amssymb}
\usepackage{graphicx}
\usepackage{dcolumn}
\usepackage{bm}
\usepackage{amscd,amsthm}
\usepackage{ifthen}
\usepackage{mathtools}
\usepackage{graphicx}
\usepackage{caption}
\usepackage{subcaption}
\usepackage{pgfplots}
\usetikzlibrary{calc,shadows}
\usetikzlibrary{pgfplots.groupplots}
\usepackage{url}
\usetikzlibrary{external}
\usepackage{amsthm}
\usepackage{mathtools}

\usepackage[nosort]{cite}        
\usepackage{url} 
\usepackage{bbm}
\usepackage{paralist}
\usepackage{caption}
\usepackage[printonlyused]{acronym}
\usepackage{xcolor}

\usepackage{times}
\usepackage{epsfig}
\usepackage{color,soul}
\usepackage{enumitem}
\usepackage{siunitx}
\usepackage{tabularx}

\newlength{\figwidth}
\pgfplotsset{every non boxed x axis/.append style={x axis line style=->,>=latex},every non boxed y axis/.append style={y axis line style=->,>=latex}}

\include{math-macros}
\include{defs}
\usepackage{hyperref}

\hypersetup{
    unicode=false,          
    pdftoolbar=true,        
    pdfmenubar=true,        
    pdffitwindow=false,     
    pdfstartview={FitH},    
    pdfauthor={Veniamin Morgenshtern},     
    pdfcreator={Veniamin Morgenshtern},   
    pdfproducer={Producer}, 
    pdfnewwindow=true,      
    colorlinks=true,       
     linkcolor=blue!60!black,          
    citecolor=blue!60!black,        
    filecolor=blue!60!black,      
    urlcolor=blue!60!black,
     hypertexnames=false
}
\usepackage{autonum}
 
\begin{document}
    
\title{Region Segmentation via Deep Learning and Convex Optimization}

\pdfinfo{
     /Title        (Region Segmentation via Deep Learning and Convex Optimization)
     /Author        (Veniamin~I.~Morgenshtern and Emmanuel J. Cand{\`e}s)
     /Keywords    ()
}

\author{
\parbox{\linewidth}{\centering
Matthias Sonntag and Veniamin I. Morgenshtern\\\vspace{0.3cm}
Chair of Multimedia Communications and Signal Processing\\
Friedrich-Alexander-Universität Erlangen-Nürnberg,\\
Erlangen 91058, Germany\\
E-mail: matthias.sonntag@fau.de, veniamin.morgenshtern@fau.de
}
}
\newcommand{\vm}[1]{{\color{red}VM: #1}}
\newcommand{\vmn}[2]{{\color{red}VM: #1} {\color{cyan} #2}}
\newcommand{\stopped}{{\color{green} STOPPED HERE!}}
\newcommand{\msn}[2]{{\color{blue}MS: #1} {\color{cyan} #2}}
\newcommand{\knn}{k}

\maketitle
\begin{abstract}
In this paper, we propose a method to segment regions in three-dimensional point clouds. We assume that (i) the shape and the number of regions in the point cloud are not known and (ii) the point cloud may be noisy. The method consists of two steps. In the first step we use a deep neural network to predict the probability that a pair of small patches from the point cloud belongs to the same region. In the second step, we use a convex-optimization based method to improve the predictions of the network by enforcing consistency constraints. We evaluate the accuracy of our method on a custom dataset of convex polyhedra, where the regions correspond to the faces of the polyhedra. The method can be seen as a robust and flexible alternative to the famous region growing segmentation algorithm. All reported results are reproducible and come with easy to use code that could serve as a baseline for future research.\footnote[1]{\url{https://github.com/vmorgenshtern/deepsegmentation}} 
\end{abstract}

\section{Introduction}
Object segmentation is one of the key problems in computer vision. Light detection and ranging (Lidar) systems are now widely used in robotics and in the automotive industry. These devices produce not images, but 3D point clouds at the output. Therefore, fast and reliable algorithms that process 3D point clouds are very important. This work is inspired by a classical algorithm, called region growing segmentation (RGS). The goal is to separate individual regions in a point cloud. When the point cloud consists of near-flat surfaces, the regions that need to be segmented are the surfaces, called \textit{faces} in the rest of the paper. For example, a cube has six faces. As we will discuss below, RGS is a greedy algorithm and thus fragile, especially when the locations of the points is noisy. We propose a two-step approach that alleviates this problem. In the following subsections, we first introduce RGS. Afterwards, we review related deep learning approaches to 3D point cloud segmentation.

\subsection{Region Growing Segmentation}
The RGS algorithm for 3D point clouds was proposed in~\cite{Rabbani.2006} and extended in~\cite{Rusu.2009}. Suppose, there is a point cloud $\mathcal{P}$ with $P$ points and each point is denoted as $\mathbf{p}_i = [x_i, y_i, z_i]^T \in \mathbb{R}^3$ with $i\in[0,...,P-1]$. The algorithm consists of four steps. (i) For each point $\mathbf{p}_i$, find the set of points $\mathcal{P}_i$ in the local neighborhood of $\mathbf{p}_i$. For example, this can be done via the $k$-nearest neighbors (KNN) algorithm.
Given some metric and a point $\mathbf{p}_i$, KNN finds the $\knn \in \mathbb{N}$ points that are closest to $\mathbf{p}_i$. This can be done efficiently in $\mathbb{R}^3$ by storing the points in a $k$-d tree data structure.
(ii) For all local neighborhoods $\mathcal{P}_i$, fit a plane and estimate normal vector $\mathbf{n}_i \in \mathbb{R}^3$ and surface curvature $\gamma_i \in \mathbb{R}$ as features for $\mathbf{p}_i$. This can be done by principal component analysis (PCA)~\cite{Rusu.2009}. (iii) Fix a threshold angle $\alpha_{th}$ in radians and threshold surface curvature $\gamma_{th}$. (These thresholds are global parameters of the algorithm that make it hard to tune.) Initially the points from $\mathcal{P}$ are assigned to different regions as follows. Initialize an empty list of seeds $\mathcal{S}$ and an empty cluster $\mathcal{C}$, add the point with minimum surface curvature to $\mathcal{S}$ and to $\mathcal{C}$. Then repeat the following procedure until $\mathcal{S}$ is empty. Choose a point $\mathbf{p}_i$ from $\mathcal{S}$. From all points $\mathbf{p}_{i_1},\ldots, \mathbf{p}_{i_\knn}$ in the local neighborhood $\mathcal{P}_i$ of $\mathbf{p}_i$, add those points to $\mathcal{C}$, that fulfill the constraint
\begin{equation}\label{angleThreshold}
\arccos(\langle \mathbf{n}_i, \mathbf{n}_{i_j}\rangle) \leq \alpha_{th},
\end{equation}
where $\langle \cdot \rangle$ denotes the inner product between two vectors and $\mathbf{n}_{i_j}$ denotes the estimated normal vector corresponding to point $\mathbf{p}_{i_j}$. The points that are added to $\mathcal{C}$ are removed from $\mathcal{P}$. If a point fulfills \eqref{angleThreshold} and  has a low surface curvature, i.e. $\gamma_{i_j} \leq \gamma_{th}$, it is added to $\mathcal{S}$. (iv) If only a low number of points is returned in $\mathcal{C}$, one may consider these points as outliers. Otherwise the current region is completed. Continue (iii) with a new region, until $\mathcal{P}$ is empty, i.e. all points in the point cloud are assigned to a region. 

One advantage of this algorithm is that it can be applied to arbitrary point clouds, as long as points are only sampled from the \emph{surface} of an object. This means that the local neighborhood is approximately flat at almost every point. An example is the surface of a cube. Except for points close to the edges, the local neighborhoods are perfectly flat. For an example where the method does not work, consider a cube filled with points. Here, it would not be possible to fit a plane to the local neighborhood of a point. Another advantage is, that there are relatively few parameters to adjust, mainly $\knn$, $\alpha_{th}$ and $\gamma_{th}$. 

There are also major drawbacks. One results from the greedy procedure of growing a region until no more points satisfy~\eqref{angleThreshold}. This means that only one erroneous connection between two individual regions results in all points of both regions being merged. Especially, if the location of points is noisy, the algorithm likely does not return satisfactory results. The reason is that the feature estimation is less accurate in noisy point clouds, which may lead to erroneously merging faces or finding too many outliers. Another disadvantage is that the parameters $\knn$, $\alpha_{th}$ and $\gamma_{th}$ need to be tuned manually and the algorithm is sensitive to the sub-optimal tuning. The parameters that work well for one point cloud will not work well for another one.

\subsection{Related approaches}
To allow similar applicability as the RGS algorithm, we identified the following requirements to the deep learning approach. The approach shall be independent of the number of individual regions in the point cloud and arbitrary shapes of the individual regions shall be allowed, as long as they are approximately flat in some direction. These requirements raise two problems. One is called the output dimension mismatch problem~\cite{Chen.2017}. In our context it means, that the number of regions and number of points vary between different point clouds. To understand the difficulty, contrast the present case with the typical supervised learning setup, say the MNIST problem, where the number of classes is fixed to~$10$ and is the same in the training set and in the test set. The second problem is called the label permutation problem~\cite{Yu.2017}. It means that the order of the individual regions is arbitrary. There is no such thing as region number one, region number two, etc; the regions are either the same or are different, but no individual labels are attached to them. For both reasons, a labeling scheme based on the individual regions is not an applicable option.

These problems also arise in other scientific domains. For segmenting different instances of objects in images, the authors of~\cite{Brabandere.2017} propose a special loss function that transforms pixels into a high dimensional embedding space. There, pixels of different object instances form clusters. A similar approach is taken in~\cite{Hershey.2016} to segment different speakers in monaural audio signals. Here, the time-frequency bins of the mixture spectrogram are transformed into a high dimensional embedding space. 

Several point cloud segmentation approaches have been proposed so far. The first approach is to substitute PCA with an improved feature estimation procedure. Robustly estimated features may alleviate some of the drawbacks of the greedy RGS approach. Deep networks that improve the normal vector estimation have been proposed in~\cite{Ben-Shabat.2019} and~\cite{Guerrero.2018}. The authors report an improvement regarding the estimated direction of the normal vectors. While this is notable, it does not remove the central problem: the greedy nature of RGS and the implied fragility. The approaches reviewed next aim at segmenting the point cloud without an explicit feature estimation step. The second approach is to project the 3D point cloud to 2D images, apply standard image segmentation techniques on the images and lift the result back to 3D. This is proposed in~\cite{Lawin.2017} and~\cite{Boulch.2017}. The significant drawback of these methods is that information is lost when projecting a 3D object to 2D. To alleviate this problem, the authors use multiple, potentially overlapping views on the point cloud. Fusing the aggregated information before backprojection to 3D is a difficult and time-consuming post-processing step. Also some points may not be visible in any of the 2D views and thus not processed. The third approach is to operate directly in 3D, but in a discrete and organized space, called the voxel space. This is different from point clouds, where points are unorganized and theoretically located in continuous space. Networks that operate this way are called VoxelNet~\cite{Huang.2016}, OctNet~\cite{Riegler.2017} and 3D U-Net~\cite{Cicek.2016}. The voxel space is constructed by small, typically cubic entities, called voxels. One may understand this as the 3D extension of pixels in images. With this space being organized, the authors of these papers propose to apply 3D extensions of convolutional neural networks. The fourth approach is to work directly on the point cloud as is done in PointNet~\cite{Qi.2017} and its extension PointNet++~\cite{Qi.2017_2}. Compared to previous methods, these methods do not require a transformation to a different space. All of the methods in the last three approaches have one drawback in common: they are only capable of segmenting labeled pre-known objects, in other words, they cannot solve the label permutation problem.

Our contributions are the following:
\begin{enumerate}[label=(\alph*)]
    \itemsep-0.3em
\item A deep learning pipeline to segment individual approximately flat regions in 3D point clouds. These regions are in the following called faces.
\item A network model that, similarly to the RGS algorithm, does not rely on the knowledge of labeled shapes of the faces or the number of faces.
\item A segmentation approach that, unlike the RGS algorithm, detects individual faces, even when there is a smooth transition between them.
\item A global, non-greedy segmentation approach.
\end{enumerate}

The rest of the paper is organized as follows. In Section~\ref{data basis}, we present the custom dataset that we used to evaluate the accuracy of our deep learning approach. Our deep learning pipeline is described in Section~\ref{methods}. Our segmentation results are presented in Section~\ref{results}. In Sections~\ref{Benchmarks}, we provide benchmarks for the runtime of the pipeline. We list interesting directions for future research in Section~\ref{Outlook} and conclude in Section~\ref{conclusion}.

\section{Custom dataset of polyhedra}
\label{data basis}
For evaluating the accuracy of our method, we require a dataset of point clouds. The point clouds need to have faces and individual faces need to be labeled. We did not find a suitable dataset, so we made a custom one. Our dataset is automatically generated from the vertices of a set of convex polyhedra. Figure~\ref{Compare} displays some of the point clouds from the dataset. We chose polyhedra, because they have a regular structure and are defined via their faces and vertices. The faces of convex polyhedra lie on their convex hull and can be derived only given the vertices. With a convex hull algorithm, the point cloud can be generated. We sample the coordinates of the points on the convex hull uniformly.

Given the vertex coordinates of different basic polyhedra, we can vary several parameters to make the dataset much richer, as described next. First, the number of points and orientation of the point cloud can be varied. Second, the point cloud can be stretched by multiplying the~$x$ and~$y$ coordinate of each point by a scalar~$\mu$ and~$\nu$. Third, the edges of the polyhedra can be rounded by iterating over all points and assigning their new location as the averaged coordinates of points in their local neighborhood. This is demonstrated in Figure~\ref{Smooth}. Fourth, normally distributed random noise $\mathbf{z}_i \sim \mathcal{N}(0, \sigma)$ with zero mean and standard deviation $\sigma$ can be added to each point $\mathbf{p}_i$. All our point clouds are normalized to just fit inside the box $[0,1]^3$ before adding the noise. To judge the magnitude of the noise in the point coordinates, $\sigma$ needs to be compared to one. 

\section{Segmentation pipeline}
\label{methods}
\begin{figure*}[t]
\begin{center}
\includegraphics[width=1.0\linewidth]{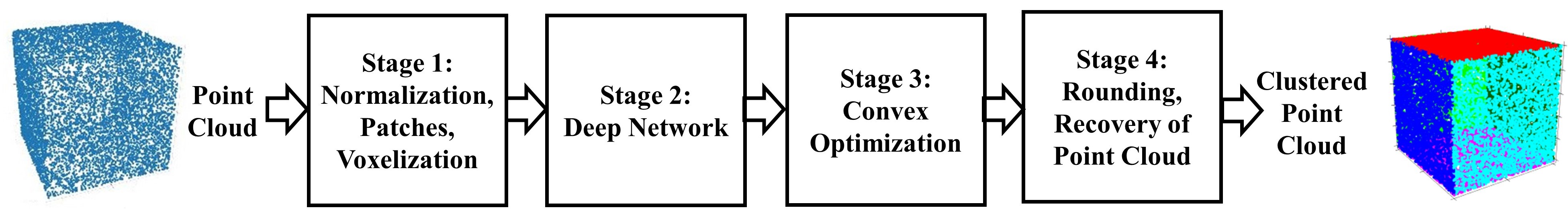}
\end{center}
   \caption{The segmentation pipeline.}
\label{overview}
\end{figure*}

In this section, we describe our proposed deep learning based approach to 3D point cloud segmentation. We begin with an overview of the whole pipeline, illustrated in Figure~\ref{overview}, and detail the components in the following subsections.
Stage 1 of the pipeline normalizes the point cloud to just fit within the range $[0,1]^3$, builds local neighborhoods and voxelizes them. We will refer to the local neighborhoods as \textit{patches}. Stage 2 of the pipeline then processes pairs of patches by a deep neural network. The network predicts the probability of the event that the two patches belong to the same face. This avoids the label permutation problem, as the exact face is not assigned at this point. This further avoids the output dimension mismatch problem, as any number of patches can be processed sequentially, yielding a defined scalar output (the probability that the pair of patches belongs to the same face). The estimated probabilities for all combinations of patches may be thresholded to obtain binary decisions that are then handed over to Stage 3. There, convex optimization is used to enforce a consistent assignment of patches. Stage 4 is a special rounding procedure, leading to the segmented point cloud.

\subsection{Stage 1: building patches of points}
\label{Preprocessing}
In this section, we describe the preprocessing stage. A point cloud is first normalized, then local patches of points are formed and those are voxelized. After the point cloud is aligned with the origin in $\mathbb{R}_+^3$, the normalization 
\begin{equation}
\tilde{\mathbf{p}}_i = \mathbf{p}_i \; \; / \max_{j,k,l \; \in \; [0,...,P-1]}\max(x_j,y_k,z_l)
\end{equation}
for $i\in[0,...,P-1]$ scales the coordinates of all points to guarantee that $\tilde{\mathbf{p}}_i\in [0,1]^3$, i.e. the normalized point cloud fits into a unit box.
This normalization does not change the proportions of the point cloud. To avoid heavy notation, from here on we will use $\mathbf{p}_i$ (not $\tilde{\mathbf{p}}_i$) to denote the points of the normalized point cloud.

Next, a set $\mathbb{P}$ that contains $N$ patches $\mathcal{P}_0,\ldots, \mathcal{P}_{N-1}$ is formed as follows. (i) A seed point $\mathbf{p}$ is chosen randomly. (ii) Together with all points in its local neighborhood, it forms a patch $\mathcal{P}$. The local neighborhood is obtained by a search method, e.g. KNN. To limit how far a patch may span into a different face, all points outside the volume of a cubic bounding box with side length $l_b$ and centered around $\mathbf{p}$ are not included in the patch. For training, the face index number of the majority of the points in $\mathcal{P}$ determines the ground truth face index number of the whole patch. (iii) The centroid of the patch 
\begin{equation} \label{centroid}
\mathbf{c} = \frac{1}{|\mathcal{P}|} \sum_{\mathbf{p} \in \mathcal{P}}\mathbf{p}
\end{equation}
is stored as a feature of $\mathcal{P}$.  In \eqref{centroid}, $| \cdot |$ denotes the cardinality of a set. After this, all points of $\mathcal{P}$ are shifted, so that they are centered at the origin:
\begin{equation} \label{centroid2}
\mathbf{p}' = \mathbf{p} - \mathbf{c}\ \text{ for all }\ \mathbf{p}\in \mathcal{P}.
\end{equation}
(iv) Repeat (i) to (iii) with a new randomly chosen seed point that is not yet part of a patch until all points belong to a patch.

\begin{figure}
\begin{center}
\includegraphics[width=1.0\linewidth]{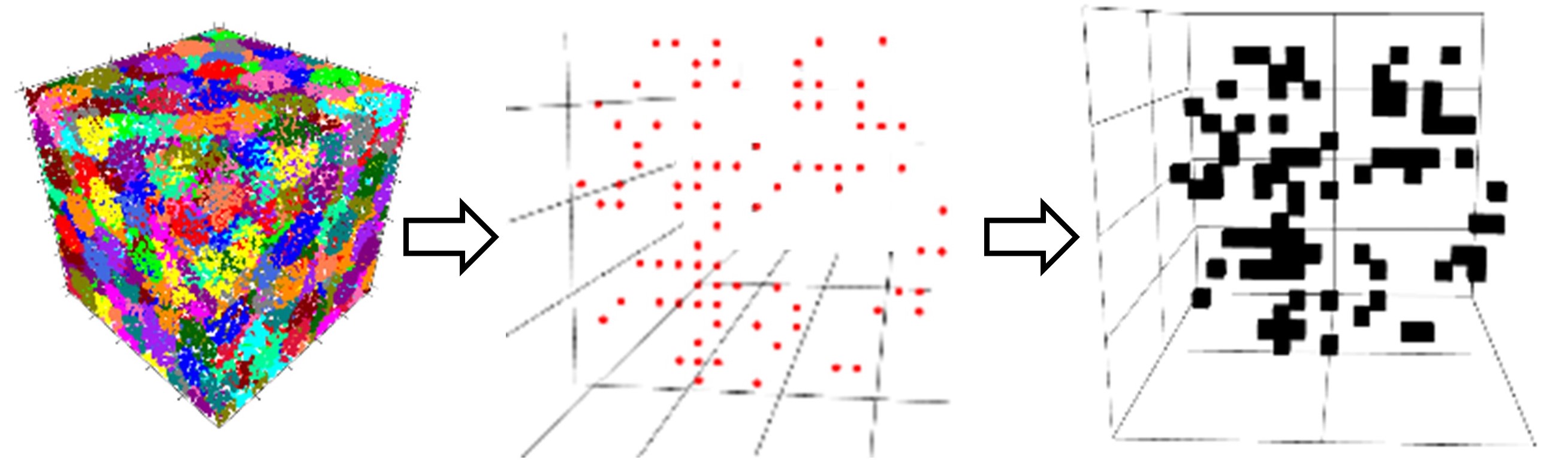}
\end{center}
   \caption{The voxelization procedure. Shown are the patches of a cubic point cloud (left), one patch of points (middle) and the voxelized patch (right). Note that colors for individual patches are reused cyclically.}
\label{Voxelize}
\end{figure}

In the last preprocessing step, all patches (centered around the origin according to~\eqref{centroid2}) in $\mathbb{P}$ are voxelized, as illustrated in Figure~\ref{Voxelize}. We use $\mathbb{V}$ to denote the set of $N$ voxelized patches $\mathcal{V}_0, \ldots, \mathcal{V}_{N-1}$. A voxelized patch is an ordered set of $M^3$ voxels, i.e. a 3D space is filled with M discrete cubes of length $l_v = l_b/M$ in each of its three dimensions. We define the origin as the center of the 3D space. The value of the voxels is binary, i.e. $\mathcal{V}_i \in \{0,1\}^{M \times M \times M}$. In accordance with~\cite{Huang.2016}, we use occupancy value, i.e. the value of a voxel is 1, if at least one point is inside the volume of the voxel. Note that unlike in the approach of \cite{Huang.2016, Riegler.2017, Cicek.2016}, we only voxelize the point cloud \emph{locally} in a small region around each patch. This way, we save computational cost by avoiding to process mostly empty space, which would happen if we would voxelize the entire point cloud globally.

\subsection{Stage 2: network to compare patches}
\label{model}
\begin{figure}[ht]
\begin{center}
\includegraphics[width=1.0\linewidth]{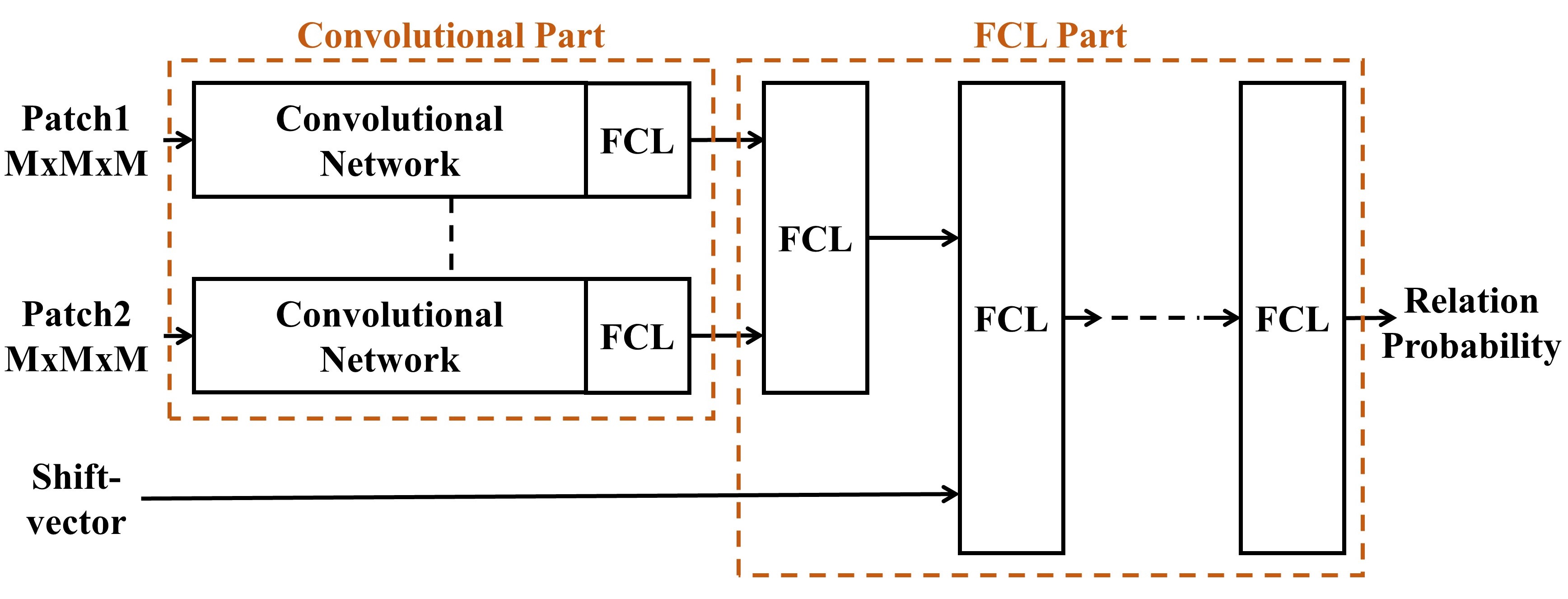}
\end{center}
   \caption{The network architecture. Input: two voxelized patches and an offset vector between the patches. Output: probability that the two patches belong to the same face.}
\label{NetworkArchitecture}
\end{figure}

In this section we first present our network model. Then we give information on how we trained it. One pass through the network computes the probability of the event that two patches belong to the same face. The idea is that the network might learn to recognize similarly oriented patches, among other hints. As patches are centered around the origin, patches from parallel faces would appear similar. To give the network a chance to distinguish such patches and determine that they belong to different faces, we also provide the relative shift vector $\mathbf{s} =  \mathbf{c}_i - \mathbf{c}_j$, where $\mathbf{c}_i$ and $\mathbf{c}_j$ are the centroids of the patches $i$ and $j$. A diagram describing the network architecture is given in Figure~\ref{NetworkArchitecture} and the details are provided in Table~\ref{OurNetwork}. 
The two patches are processed individually by the same 3D Convolutional Neural Network (CNN). 3D convolutions take a 4D input tensor ($x$ voxels $\times$ $y$ voxels $\times$ $z$ voxels $\times$ features) and produce a 4D output tensor by convolving in $x$, $y$, and $z$ directions and fully-connecting in feature direction. We tried a 3D variant of the AlexNet~\cite{Krizhevsky.2012} and Residual Networks (ResNets)~\cite{He.2016}. Only with the ResNet, we observed satisfactory results. The key component of ResNets are the residual blocks. Let $\mathbf{x}$ denote the input tensor to the residual block, then the output is given by
\begin{equation} \label{ResNetMap}
\mathbf{y} = \text{ReLU}(\mathcal{F}(\mathbf{x}) + \mathbf{x}),
\end{equation}
where~$\mathcal{F}(\mathbf{x})$ is obtained from~$\mathbf{x}$ by sequentially applying Conv3D -- BatchNorm3D -- ReLU -- Conv3D -- BatchNorm3D. Conv3D is a 3D convolutional layer, BatchNorm3D denotes the 3D variant of batch normalization~\cite{Ioffe.2015} and ReLU is the rectified linear unit. The idea is that the identity mapping of the input~$\mathbf{x}$ is added to the learned mapping~$\mathcal{F}(\mathbf{x})$ just before the output of the block. The authors of~\cite{He.2016} showed that training very deep networks is possible by stacking these residual blocks. 

After the two patches are processed by the convolutional network, the individual results are summarized by a fully connected layer (FCL) and combined with another FCL. This output is then combined with the shift vector~$\mathbf{s}$ and processed by several FCLs. Applying the softmax function to the output of the last layer yields the probability that two patches belong to the same face. As activation function, we use ReLU throughout the network. We initialize all weights and biases with Kaiming initialization~\cite{He.2015}. 
\begin{table}
\begin{center}
\begin{tabular}{l|l}
\multicolumn{1}{c}{\bf Convolutional part}  &\multicolumn{1}{c}{\bf FCL part}\\ \hline
Conv3D(1, 8, 3, 1, 1) & \\
BatchNorm3D, ReLU & \\ 
MaxPool3D(8, 8, 3, 1, 1)    & FCL(256, 30), ReLU \\ 
ResBlock3D(8, 8, 3, 1, 1)   & FCL(33, 30), ReLU \\
ResBlock3D(8, 16, 3, 2, 3)  & FCL(30, 30), ReLU \\
ResBlock3D(16, 32, 3, 2, 3) & FCL(30, 20), ReLU \\
ResBlock3D(32, 64, 3, 2, 3) & FCL(20, 2), Softmax\\
ResBlock3D(64, 128, 3, 2, 3) &\\
Global Average Pool(128, 128) &\\
FCL(128, 128), ReLU
\end{tabular}
\end{center}
\caption{The details of our network architecture. Left: Convolutional part of the network for one patch (identical for second patch), right: FCL part of the network. Notation: Conv3D (in feat, out feat, kernel size, stride, padding), MaxPool3D (in feat, out feat, kernel size, padding), ResBlock3D (in feat, out feat, kernel size, stride, padding), Global Average Pool (in feat, out), FCL (in, out). We pad with zeros at the borders. Kernels have the same size in all dimensions. All ResBlock3D are constructed according to~\eqref{ResNetMap}. The second Conv3D layer in a ResBlock3D uses the same kernel size as the first one with a stride of 1 and padding of 1.}
\label{OurNetwork}
\end{table}
For training, we used the weighted cross-entropy loss function as implemented in PyTorch~\cite{Pytorch.2017}. As optimizer, we tried mini-batch gradient descent~\cite{Ruder.2016} and Adam~\cite{Kingma.2014}. With the latter, we observed better generalization.
For an update step, we fix one random patch and consider~$N$ pairs between this patch and all other patches in the point cloud. The loss for this step is the average loss over these~$N$ pairs. For each point cloud we repeat this update step $L=\max(N, 50)$ times, fixing a new random patch every time. This way to process one point cloud we need to do $N\cdot L$ forward and backward passes through the network. This is much faster than doing a forward and backward pass $N^2$ times, corresponding to all combinations of patches in the point cloud. In one epoch we process all point clouds in the training set in this way.

\subsection{Stage 3: consistency via convex optimization}
\label{convexOp}
In this section, we explain how the (possibly noisy) pairwise probabilities that the patches in the pair belong to the same face, can be made \emph{globally} consistent using convex optimization.

At the input of stage 3, we have a binary matrix $\mathbf{X}_{hard}^{in}~\in~\{0,1\}^{N \times N}$ or real matrix $\mathbf{X}_{soft}^{in}~\in~\mathbb{R}_+^{N \times N}$. The entry~$(i,j)$ of the binary matrix is obtained by thresholding the probability of the event that patches~$i$ and~$j$ belong to the same face, as predicted by the network in stage 2 of the pipeline; the threshold we use is~$0.5$. To simplify notation, we use $\mathbf{X}^{in}$ in the following for both cases.
The network has no global information, so~$\mathbf{X}^{in}$ is likely inconsistent. The next step is to denoise~$\mathbf{X}^{in}$ and find a consistent matrix $\mathbf{X}~\in~\mathbb{R}^{N \times N}$. It turns out that a closely related problem has been studied~\cite{Chen.2014}. The authors developed the MatchLift algorithm that allows to find correspondences between multiple somewhat different views of the same object. Our contribution is to translate the MatchLift algorithm to our setting and use it for denoising the matrix~$\mathbf{X}^{in}$. This can be done using the following considerations.

Suppose there are $m$ faces. Let $\mathbf{Y} = \{0,1\}^{N \times m}$ denote the matrix in which the $(i,j)$th element is one, iff the patch number $i$ belongs to the face number $j$. For example, in the case of a cube, in which each face consists of two patches, the matrix $\mathbf{Y}\in\{0,1\}^{12 \times 6}$ is 
\begin{equation}
    \mathbf{Y} =
    \begin{bmatrix}
        1 & 0 & 0 & 0 & 0 & 0 \\
        1 & 0 & 0 & 0 & 0 & 0 \\
        0 & 1 & 0 & 0 & 0 & 0 \\
        0 & 1 & 0 & 0 & 0 & 0 \\
        \vdots & \vdots & \vdots & \vdots & \vdots & \vdots\\
        0 & 0 & 0 & 0 & 0 & 1 \\
        0 & 0 & 0 & 0 & 0 & 1
    \end{bmatrix}
\end{equation}
We can see that the ideal $\mathbf{X}$ can be obtained as $\mathbf{X}~=~\mathbf{Y}\mathbf{Y}^T$. If follows that $\mathbf{X}$ is symmetric, positive semidefinite (PSD), contains~$1$ on the diagonal, has binary values, and $\mathrm{rank}(\mathbf{X})\!=\!m$.
Therefore, to recover $\mathbf{X}$ from $\mathbf{X}^{in}$ it is natural to solve the following optimization problem:
\begin{equation}\label{Opt}
\begin{aligned}
& \underset{\mathbf{X}}{\text{maximize}}
& & \langle \mathbf{X}, \mathbf{X}^{in} \rangle \\
& \text{subject to}
& & \mathrm{rank}(\mathbf{X}) = m, \\
& & & 0\le \mathbf{X}_{ij} \le 1, \mathbf{X} \succeq 0, \mathbf{X}_{ii} = 1.
\end{aligned}
\end{equation}
Intuitively the optimization tries to find a matrix that is well correlated with the input data, subject to all the constraints applicable to $\mathbf{X}$.
Unfortunately, due to the $\mathrm{rank}(\mathbf{X}) = m$ constraint this problem is nonconvex and there is no efficient way to solve it. As explained in~\cite{Chen.2014}, a good convex surrogate for~\eqref{Opt} is  
\begin{equation}\label{convOpt}
\begin{aligned}
& \underset{\mathbf{X}}{\text{minimize}}
& & -\langle \mathbf{X}, \mathbf{X}^{in} \rangle + \frac{1}{2} \langle \mathbf{X}, \mathbf{1}\mathbf{1}^T \rangle \\
& \text{subject to}
& & \mathbf{X} \geq 0, \begin{bmatrix}
m & \mathbf{1}^T\\
\mathbf{1} & \mathbf{X}\\
\end{bmatrix} \succeq 0, \mathbf{X}_{ii} = 1,
\end{aligned}
\end{equation}
where $\mathbf{1}$ denotes a vector of ones.
In practice we often do not know $m$. A method is proposed in~\cite{Chen.2014} that allows to estimate $m$ based on the eigenvalues of~$\mathbf{X}^{in}$. We found this method to work poorly in our problem, because if $m$ is underestimated, then the optimization problem in~\eqref{convOpt} will return a matrix that corresponds to a point cloud with a lower number of faces than the one in the correct clustering; unrelated clusters will be joined. This is a failure mode we would like to avoid. Hence, we found that taking a conservative empirical overestimate for $m$ works best.

\subsection{Stage 4: rounding procedure}
\label{rounding}
A special rounding procedure is applied to the matrix $\mathbf{X}$ and directly yields the required clusters. 
The rounding procedure suggested in \cite{Chen.2014} is not transferable to our case, so we designed a new procedure that works as follows.
We fix one patch assignment after the other by correlating the columns in a temporary matrix~$\mathbf{A}$. The matrix $\mathbf{A}$ is initialized as~$\mathbf{A} = \mathbf{X}$. A list of lists $\mathcal{C}=[C_1,C_2, \ldots]$ is used to allocate patch indices to clusters: if $j\in C_i$, then the patch~$j$ belongs to the cluster~$i$. It is initialized as $\mathcal{C}~=~[[0], [1], ... , [N-1]]$, where~$N$ is the number of patches. Then the following steps lead to a set of clusters. (i) We compute $\mathbf{A}^T\mathbf{A}$ and find its maximum entry outside the main diagonal, or in other words, we find two distinct columns of $\mathbf{A}$ with the largest inner product.
If the maximum entry, called $(i, j)$ with slight abuse of notation, exceeds a threshold,
\begin{equation}\label{thresh}
\underset{i,j \land i \neq j}{\max}[\mathbf{A}^T \mathbf{A}]_{i,j} \geq \frac{n}{2},\quad \text{where}\quad~n = N/m,
\end{equation}
then the $j$th list in $\mathcal{C}$ is merged with its~$i$th list, i.e. patches~$i$ and~$j$ belong to the same cluster. Note that if all faces would contain the same number of patches, then each face would have $n$ patches. With $n/2$ as threshold, we allow a varying number of patches per face. This variety is limited, as small faces lead to sparse columns in $\mathbf{A}$ and thus a small inner product. (ii) Rebuild $\mathbf{A}$ from $\mathbf{X}$ in accordance with the updated $\mathcal{C}$. Each column~$i$ is constructed as follows:
\begin{equation}
\mathbf{A}_{:i} = \frac{1}{|\mathcal{C}_i|} \sum_{j=0}^{|\mathcal{C}_i|-1}\mathbf{X}_{:\mathcal{C}_i[j]}.
\end{equation}
Above, $\mathcal{C}_i[j]$ denotes the $j$th element in $\mathcal{C}_i$. This rebuilding step reduces the width of $\mathbf{A}$ by 1 in each iteration and averages the column entries of $\mathbf{X}$ such that $[\mathbf{A}^T\mathbf{A}]_{i,j} \leq N$ for all $i$ and $j$.
(iii) Repeat (i) and (ii) as long as~\eqref{thresh} cannot be fulfilled. Then all patch indices are assigned to a cluster.

The clustered point cloud is found by iterating over all list elements in the cluster list~$\mathcal{C}$ and retrieving the points via the patch indices herein.

\section{Results}
\label{results}
In this section we compare the accuracy of our approach to the classical RGS approach. An implementation of RGS is available as part of the Point Cloud Library (PCL)~\cite{Rusu.2011}. For our evaluation, we reimplemented this algorithm.
 
Our dataset is organized in three sets: a training and validation set for training and a test set for evaluation. The vertices of~$15$ polyhedra are used to generate 500 point clouds for the training set and 50 point clouds for the validation set. The vertices of three \emph{different} polyhedra (not the ones used for the training and validation sets) are used to generate a disjoint test set with 50 point clouds. By randomly varying the generator parameters as explained in Section~\ref{data basis}, we obtain diverse data. A sample of the sets together with the generator configuration can be found in the appendix. Our discussion below reveals that the classical RGS algorithm is somewhat superior to our approach for ideal point clouds: the ones with sharp edges between the faces and with no noise. When the point cloud is noisy, as is usually the case in applications, our approach is much more robust than RGS. Further, our approach does not require manual parameter tuning, while good parameters for RGS are point cloud specific. Finally, and most importantly, we show, that our approach does not merge faces that are connected by rounded edges, whereas the RGS algorithm fails in this case.

\subsection{Accuracy}
\label{Performance}
\begin{figure*}[t]
\begin{center}
\includegraphics[width=1.0\linewidth]{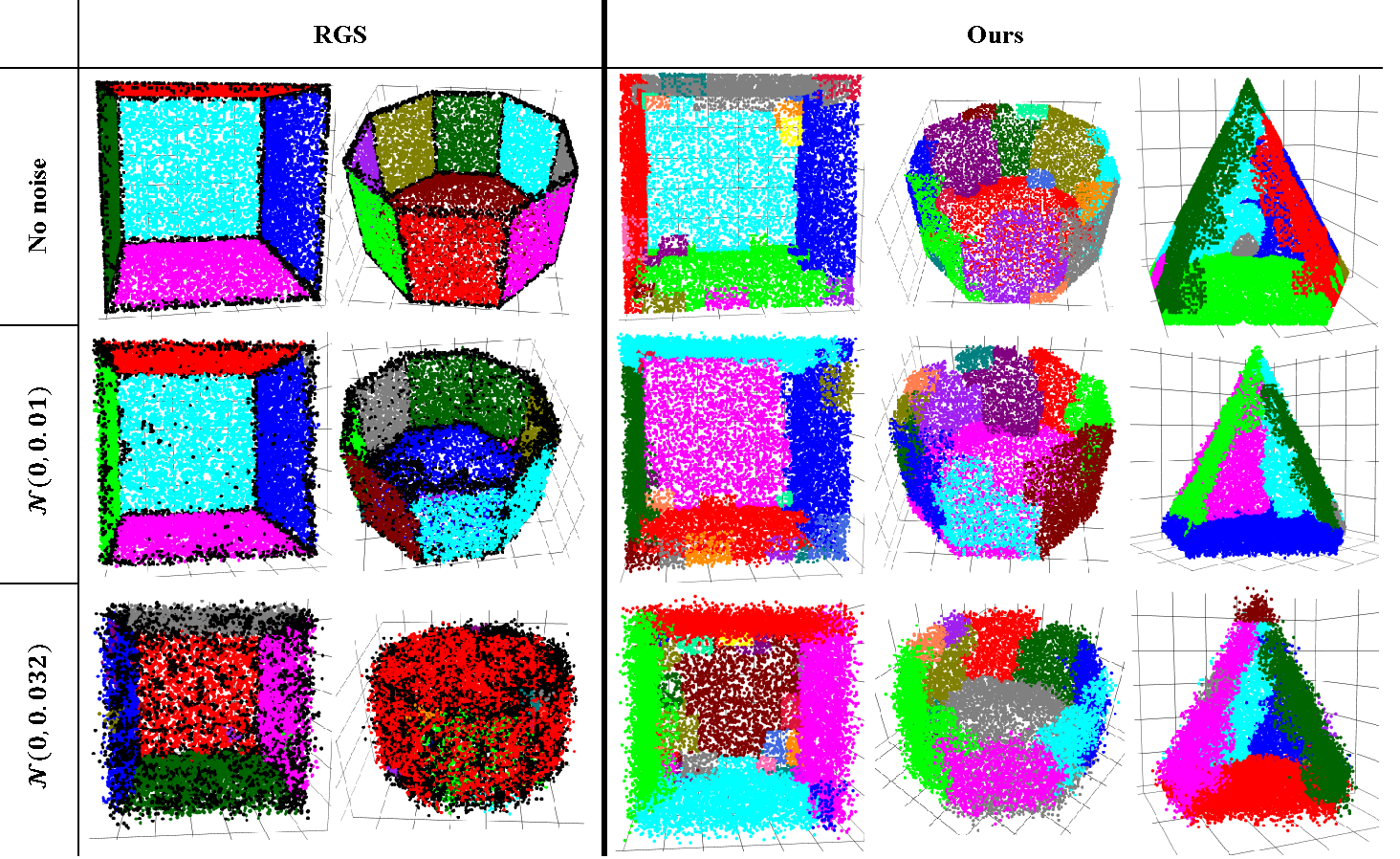}
\end{center}
\caption[.]{Comparison between clustering accuracy of RGS and our approach. Shown are a cube, octagonal prism and pentagonal pyramid. These polyhedra types are from the test set and have never been seen by the network during training. The point clouds are normalized to just fit within $[0,1]^3$. The noise level is varied. Faces showing towards the reader were made invisible. Outlier clusters are only returned by the RGS algorithm and drawn in black. Parameters for RGS: ($\knn=20$, $\alpha_{th}=3^{\circ}$, $\gamma_{th}=1.0$) for noiseless data; ($\knn=30$, $\alpha_{th}=5^{\circ}$, $\gamma_{th}=1.0$)  for $\mathcal{N}(0, 0.01)$; ($\knn=35$, $\alpha_{th}=10^{\circ}$, $\gamma_{th}=1.0$) for $\mathcal{N}(0,0.032)$. For our method, we display the point clouds as obtained when the real matrix $\mathbf{X}^{in}_{soft}$ is processed by MatchLift with $m=14$.}
\label{Compare}
\end{figure*}

Figure~\ref{Compare} shows the clustering accuracy of RGS and of our approach for different noise levels. The left two columns are related to RGS, the right three columns give examples for our approach. For RGS, it is easy to select good values for parameters~$\knn$,~$\alpha_{th}$ and~$\gamma_{th}$ for ideal (noiseless) point clouds; the algorithm performs near-perfectly. When noise is added, it becomes increasingly difficult to select good values for the parameters. Even worse: the good choice depends on the point cloud. We searched for a good parameter setting that balances the number of outlier clusters and correctly identified faces. It is visible, that RGS returned very good results for the noiseless point clouds. For low noise though, it returned many outlier clusters and did not allow a configuration that would be optimal for cube and octagonal prism simultaneously. For the octagonal prism, four faces, drawn in dark green and cyan have erroneously been merged. For higher noise, even more outlier clusters were returned for the cube. The same setting lead to failure on the octagonal prism.

In our approach, the boundaries between the faces are not as cleanly delineated as in RGS. This is because we make decision on patch level, not on point level; the patches may overlap from one face to another. We accept this degradation on the boundaries and leave it to future work to fix. There are several options to solve this problem via a \emph{local} post-processing refinement. Since, as explained in Section~\ref{Preprocessing}, the patches are restricted from extending more than $l_b/2$ around a seed point, the imprecision on the boundaries is also limited to this strict bound. In this work our focus is correct \emph{global} clustering, a more challenging problem. More on this follows in Section~\ref{connectedface}.

The accuracy of our approach is stable for all noise levels.
The robustness may be attributed to convolutional networks that learn to extract relevant features, such as the orientations of the patches, even in the presence of noise. One can see, that the faces are properly detected. Only for high noise with $\sigma\!=\!0.032$, the two light green faces on the left of the corresponding octagonal prism were erroneously merged. Local degradations are visible especially at the edges, where single patches are not merged with any face. These contain parts of several faces and are thus difficult to assign.

We experimented with real and binary matrices at the input of MatchLift, omitting MatchLift and oracle assignment to the parameter~$m$ instead of using a fixed $m=14$. Table~\ref{Performancetab} gives the pipeline accuracy on the test set. The accuracy is measured as the number of correctly identified patch combinations divided by the total number of combinations,~$N^2$. One can see that the accuracy with MatchLift is \emph{in all cases} higher than that without MatchLift. Especially remarkable is the high noise level case with standard deviation $\sigma = 0.032$: approximately~$30\%$ accuracy is gained by using MatchLift. Even in the situation when it is inherently hard to make reliable decisions based on pairs of patches, a global convex-optimization based approach is powerful enough to find a near-perfect global assignment. Using a binary or real matrix at the input of MatchLift does not lead to significantly different results. Using an oracle assignment for~$m$ compared to setting a fix value,~$m=14$, did only yield minor improvements for the highest noise level with $\sigma = 0.032$. Visually checking the clustered point clouds revealed that the global structure was identified consistently for the cubes and pentagonal prisms. For the octagonal prisms, the large top and bottom face were not consistently separated, if the gap between both faces was in the order of~$l_b$ or smaller. This happened partially, because the patches contain points from both faces.
\begin{table}[h]
\centering
\small
\begin{tabularx}{0.6\linewidth}{X|c|c|c|} 
&No noise & $\sigma\!=\!0.01$ & $\sigma\!=\!0.032$ \\
\hline
$\mathbf{X}^{in}_{hard}$, no ML, $m\!=\!14$ & $93.71\%$ & $90.53\%$ & $66.46\%$ \\
\hline
$\mathbf{X}^{in}_{soft}$, no ML, $m\!=\!14$ & $93.23\%$ & $89.40\%$ & $55.34\%$ \\
\hline
$\mathbf{X}^{in}_{hard}$, ML, $m\!=\!14$ & $95.29\%$ & $93.67\%$ & $89.16\%$ \\
\hline
$\mathbf{X}^{in}_{soft}$, ML, $m\!=\!14$ & $95.27\%$ & $93.92\%$ &$89.38\%$ \\
\hline
$\mathbf{X}^{in}_{hard}$, ML, oracle~$m$ & $95.43\%$ & $94.14\%$ & $\textbf{91.16\%}$ \\
\hline
$\mathbf{X}^{in}_{soft}$, ML, oracle~$m$ & $\textbf{95.64\%}$ & $\textbf{94.17\%}$ &$91.13\%$ \\
\hline
\end{tabularx} 
\caption{Mean accuracy in \% for different processing methods and noise levels. A normally distributed variable $\mathbf{z}~\sim~\mathcal{N}(0, \sigma)$ is added to the coordinates of the points. The point clouds are standardized to just fit within a $[0,1]^3$ box \emph{before} the noise is added. Given is the accuracy averaged over 50 point clouds in the test set. For model selection, we evaluated the accuracy of the network on the validation set after every epoch and chose the network that achieved the best average accuracy. For each noise level, we trained for 13 epochs with learning rate $\eta=10^{-4}$. In all cases, we used the Adam optimizer. With ML, we abbreviate MatchLift.}
\label{Performancetab}
\end{table}

\subsection{Segmenting connected faces}
\label{connectedface}
\begin{figure}[t]
\begin{center}
\includegraphics[width=1.0\linewidth]{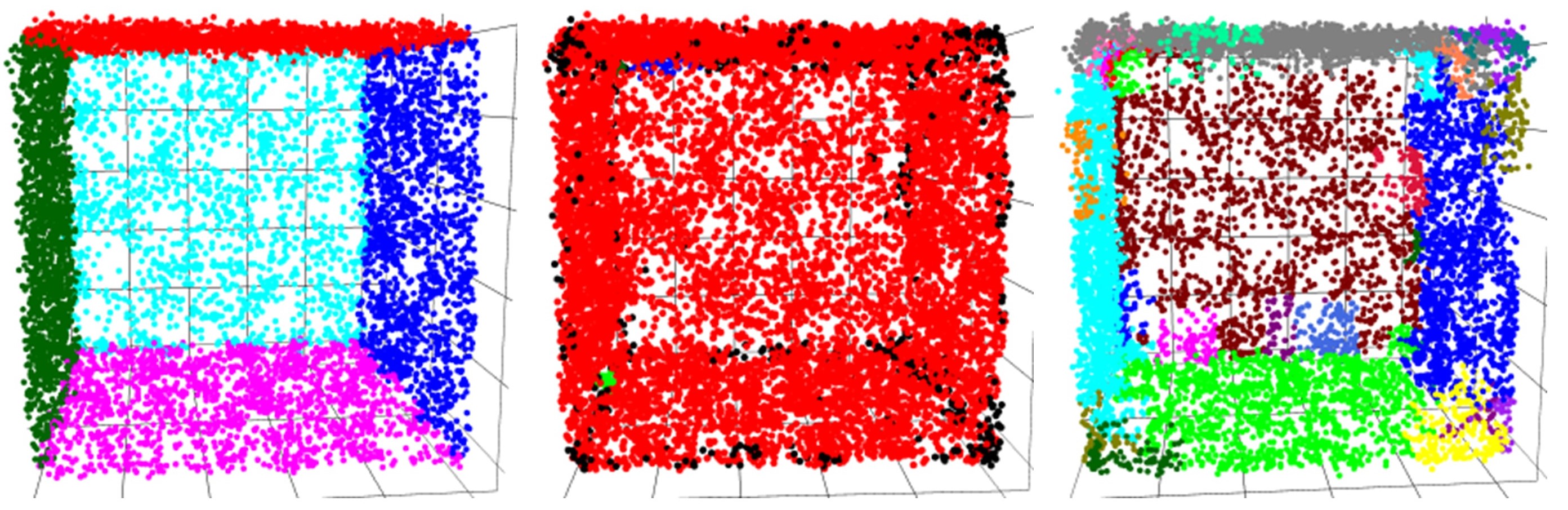}
\end{center}
   \caption{The clustering behavior of RGS and our method differs significantly for faces that are connected by a smooth or round edge. Shown are the ground truth point cloud on the left, the result of RGS ($\knn~=~20,~\alpha_{th}=10$,~$\gamma_{th}=1.0$) in the middle and the result of our method on the right. Individual faces showing towards the reader were made invisible.}
\label{Smooth}
\end{figure}

RGS greedily merges faces based on the angular difference between the normal vectors attributed to a point. Faces that are connected by a smooth or rounded edge are thus likely merged by RGS. Even worse: consider two faces that are connect by a sharp edge, but there is a single small smooth transition between the two faces. Because of its greedy nature, RGS will merge such faces. We propose a robust alternative: our approach does not merge smoothly connected faces and makes decisions based on the \emph{global} structure. To demonstrate this, we prepared a cube with smooth edges and show in Figure~\ref{Smooth} the different segmentation results. In this case we can see that RGS merges all faces, since all faces are smoothly connected. Our algorithm returns several unconnected patches at the edges, but gets the global structure correctly.

\section{Benchmarks}
\label{Benchmarks}
Here we give rough estimates on computational complexity of our pipeline for training and inference. One training epoch with 500 point clouds takes about eight hours on an Intel\textsuperscript{\textregistered} Xeon\textsuperscript{\textregistered} Silver 4114 CPU and an MSI Nvidia GeForce GTX 1080 Ti GPU. Pre- and postprocessing are not executed during training. Taking the average processing time over 50 point clouds, inference on one point cloud takes about one minute for preprocessing and evaluating all combinations of patches by the network. Convex Optimization and rounding takes about 30 seconds. For these benchmarks, each point cloud had between $5000$ and $50000$ points. Detailed information on all our settings is given in the appendix.

\section{Outlook}
\label{Outlook}
As presented so far, it may appear that our method only works for objects with nearly flat surfaces. This is not so. The case of polyhedra with near flat edges is the easiest to explain and to experiment with. The method, however, applies whenever reasonably reliable information about the relationship between the pairs of patches may be inferred from the local properties of these patches and the offset vector. For example, the local properties might rely on the information about texture of the patches, curvature of patches, etc. Depending on the problem at hand, the patches may be taken large or small. It is not necessary to voxelize the patches. Instead of the voxel representation, one can compute a local statistic for each patch that consists of just a few numbers: the normal vector, the curvature, etc. This would lead to a very fast implementation, that, however, would be insensitive to fine-grained information in the patches. The general structure of the network in Figure~\ref{NetworkArchitecture} will remain the same, but the details of the convolutional branches will change. Further, one can consider applying our approach in a semi-greedy way: (i) train a network that reliably predicts pairwise relationships about patches that are not too far away from each other, (ii) apply the network locally, leaving the relationships between patches that are further away undefined, (iii) apply convex optimization to find a globally consistent assignment. We leave it to future work to explore all these directions fully.

\section{Conclusion}
\label{conclusion}
We proposed a deep learning based method for finding individual faces in 3D point clouds. Same as the classical RGS algorithm, we only require the set of points as input and are not limited to known objects. In contrast to RGS, our method is non-greedy and uses global information. It is robust and once trained, does not require manual parameter tuning. Smoothly connected faces are not merged by our method. Our results are fully reproducible and we make the complete source code available. We also include all the trained models that were used in evaluations in this paper. 

\interlinepenalty=10000
{\small
\bibliographystyle{ieee_fullname}
\bibliography{deepsegmentation}
}

\newpage
\appendices
\section{Appendix}
\subsection{Point cloud generator configuration}
\label{genconfig}
\begin{table}[h]
\centering
\begin{tabular}{|r|c|c|c|}
\hline 
Parameter & Apply probability & Lower bound & Upper bound \\
\hline 
Number points in a point cloud & - & $5000$ & $50000$ \\
\hline 
Scaling factor & $1$ & $1$ & $1$ \\
\hline 
Neighbourhood for edge rounding $k = \frac{\text{Points~in~Point~Cloud}}{\alpha}$ & $0.5$ & $\alpha = 100$  & $\alpha=10000$ \\ 
\hline 
Roll, pitch, yaw rotation [degree]& $1$ & $0$ & $360$ \\ 
\hline 
Stretching factor $\mu$ and $\nu$ for x,y axis & $1$ & $0.5$ & $1$ \\ 
\hline 
\end{tabular} 
\caption{The configuration of the point cloud database generator. Before a point cloud is generated, each parameter is sampled uniformly from values between the upper and lower bound.}
\end{table}

\subsection{Pipeline configuration}
\label{pipelineconfig}

\begin{table}[h]
\centering
\begin{tabular}{|r|c|}
\hline
Parameter & Value \\
\hline 
Number point clouds (training set) & $500$ \\ 
\hline 
Number point clouds (validation set) & $50$ \\ 
\hline 
Number point clouds (test set) & $50$ \\ 
\hline 
Number training epochs & $13$ \\ 
\hline
Optimizer & Adam \\
\hline
Learning rate & $10^{-4}$ \\
\hline
Loss function & Cross-Entropy Loss \\
\hline
Weight ratio for loss function & $8$ \\
\hline
Number of patches compared to each patch during training, $L$ & 50 \\
\hline
Probability threshold to accept binary relation for $\mathbf{X}^{in}_{hard}$ & $0.5$ \\
\hline
Preprocessing: search method & KNN \\
\hline
Preprocessing: $k$ for KNN & $\frac{\text{Nr.~points~in~point~cloud}}{50}$ \\
\hline
Preprocessing: length $l_b$ of patch boundary cube & $0.2$ \\
\hline
Preprocessing: length voxelization box & $0.2$ \\
\hline
Preprocessing: number voxels per dimension & $21$ \\
\hline
\end{tabular}
\caption{Parameters of the segmentation pipeline.}
\end{table}
\vfill 

\newpage

\subsection{Extracts from datasets}
\label{extracts}
\begin{table}[h!]
\begin{center}
\includegraphics[width=0.8\linewidth]{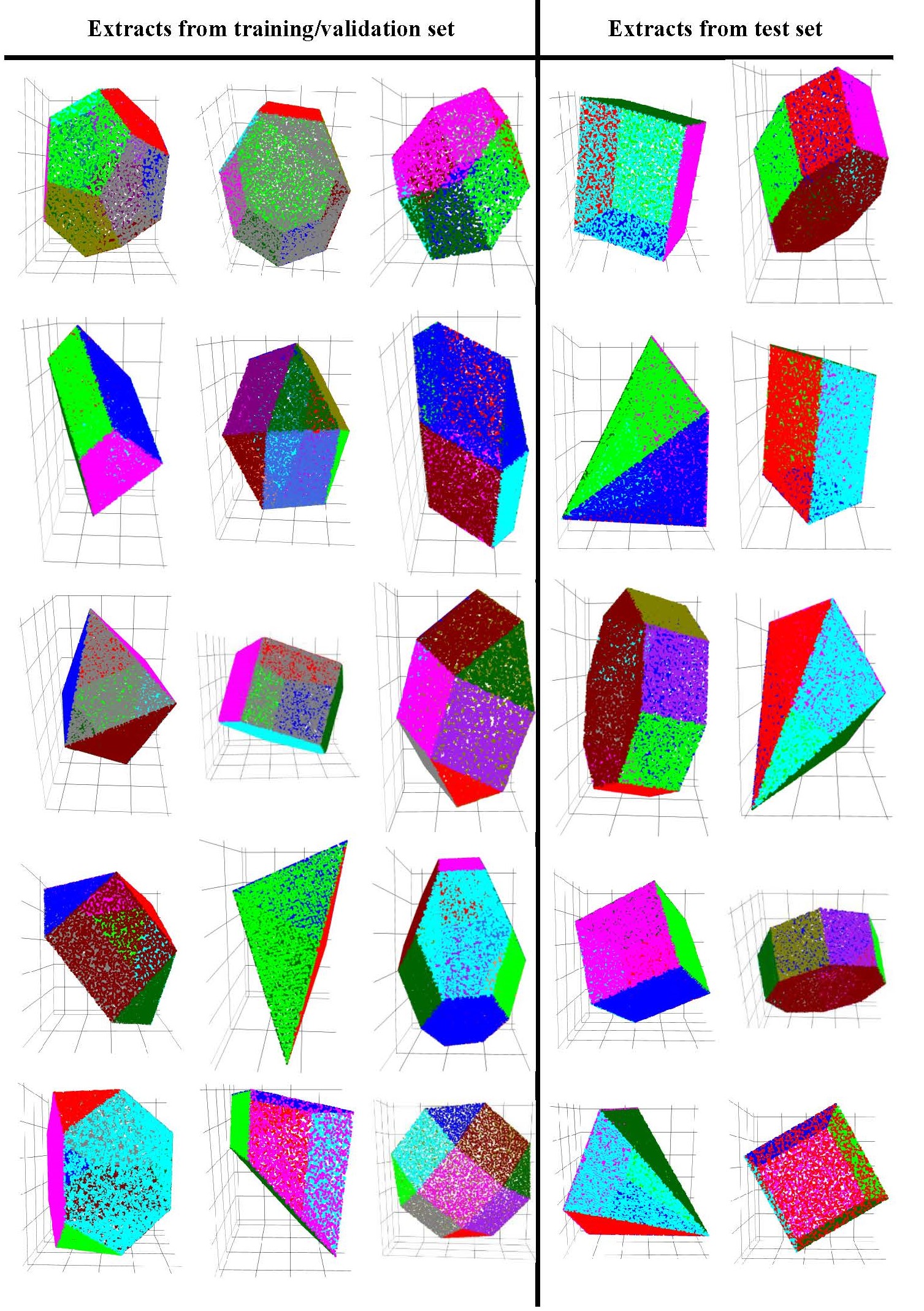}
\end{center}
   \caption{Some ground truth point clouds from the noiseless training and validation set (left) and test set (right).}
\label{network}
\end{table}

\end{document}



\title{Appendix for \\
Region Segmentation via Deep Learning and Convex Optimization}


\maketitle

\renewcommand{\arraystretch}{1.3}

\section{Point cloud generator configuration}
\label{genconfig}
\begin{table}[h]
\centering
\begin{tabular}{|r|c|c|c|}
\hline 
Parameter & Apply probability & Lower bound & Upper bound \\
\hline 
Number points in a point cloud & - & $5000$ & $50000$ \\
\hline 
Scaling factor & $1$ & $1$ & $1$ \\
\hline 
Neighbourhood for edge rounding $k = \frac{\text{Points~in~Point~Cloud}}{\alpha}$ & $0.5$ & $\alpha = 100$  & $\alpha=10000$ \\ 
\hline 
Roll, pitch, yaw rotation [degree]& $1$ & $0$ & $360$ \\ 
\hline 
Squashing factor $\mu$ and $\nu$ for x,y axis & $1$ & $0.5$ & $1$ \\ 
\hline 
\end{tabular} 
\caption{The configuration of the point cloud database generator. Before a point cloud is generated, each parameter is sampled uniformly from values between the upper and lower bound.}
\end{table}

\section{Pipeline configuration}
\label{pipelineconfig}

\begin{table}[h]
\centering
\begin{tabular}{|r|c|}
\hline
Parameter & Value \\
\hline 
Number point clouds (training set) & $500$ \\ 
\hline 
Number point clouds (validation set) & $50$ \\ 
\hline 
Number point clouds (test set) & $50$ \\ 
\hline 
Number training epochs & $13$ \\ 
\hline
Optimizer & Adam \\
\hline
Learning rate & $10^{-4}$ \\
\hline
Loss function & Cross-Entropy Loss \\
\hline
Weight ratio for loss function & $8$ \\
\hline
Number of patches compared to each patch during training, $L$ & 50 \\
\hline
Probability threshold to accept binary relation for $\mathbf{X}^{in}_{hard}$ & $0.5$ \\
\hline
Preprocessing: search method & KNN \\
\hline
Preprocessing: $k$ for KNN & $\frac{\text{Nr.~points~in~point~cloud}}{50}$ \\
\hline
Preprocessing: length $l_b$ of patch boundary cube & $0.2$ \\
\hline
Preprocessing: length voxelization box & $0.2$ \\
\hline
Preprocessing: number voxels per dimension & $21$ \\
\hline
\end{tabular}
\caption{Parameters of the segmentation pipeline.}
\end{table}

\pagebreak
\section{Extracts from datasets}
\label{extracts}
\begin{table}[h!]
\begin{center}
\includegraphics[width=0.8\linewidth]{figures/ExtractsData.jpg}
\end{center}
   \caption{Some ground truth point clouds from the noiseless training and validation set (left) and test set (right).}
\label{network}
\end{table}